\documentclass[10pt,twocolumn,letterpaper]{article}

\usepackage[pagenumbers]{cvpr} %

\usepackage{algorithm}
\usepackage{algpseudocode}
\usepackage{amsfonts}
\usepackage{amsmath}
\usepackage{booktabs}
\usepackage{enumitem}
\usepackage{graphicx}
\usepackage{hhline}
\usepackage{mathtools}
\usepackage{multirow}
\usepackage{nicefrac}
\usepackage{pifont}
\usepackage{soul}
\usepackage{tabularx}
\usepackage{booktabs}
\usepackage{array}
\usepackage{makecell}
\usepackage{url}
\usepackage{wrapfig}
\usepackage{listings}
\lstdefinestyle{snippet}{basicstyle=\ttfamily\small, breaklines=true, frame=single, columns=fullflexible}

\newcommand{\benchmark}{S2VTime\xspace}

\newcommand{\clipt}{$\textnormal{CLIP}_\textnormal{text}$\xspace}
\newcommand{\clipr}{$\textnormal{CLIP}_\textnormal{ref}$\xspace}

\newcolumntype{Y}{>{\centering\arraybackslash}X}

\definecolor{cvprblue}{rgb}{0.21,0.49,0.74}
\usepackage[pagebackref,breaklinks,colorlinks,allcolors=cvprblue]{hyperref}

\def\paperID{19225} %
\def\confName{CVPR}
\def\confYear{2026}

\newcommand{\ours}{AlcheMinT\xspace}
\newcommand{\myparagraph}[1]{\noindent\textbf{#1}}

\title{\ours: Fine-grained Temporal Control for Multi-Reference Consistent Video Generation}

\author{
Sharath Girish$^{1}$ \and Viacheslav Ivanov$^{1}$ \and Tsai\mbox{-}Shien Chen$^{1,2}$ \and Hao Chen$^{1}$ \and
Aliaksandr Siarohin$^{1}$ \and Sergey Tulyakov$^{1}$ \and \\[-5pt]
$^{1}$Snap Inc. \quad $^{2}$UC Merced \\
Project page: \href{https://snap-research.github.io/Video-AlcheMinT}{https://snap-research.github.io/Video-AlcheMinT}
}
\begin{document}

\twocolumn[{%
\maketitle
\vspace{-2em}

\begin{center}
  \includegraphics[width=\linewidth]{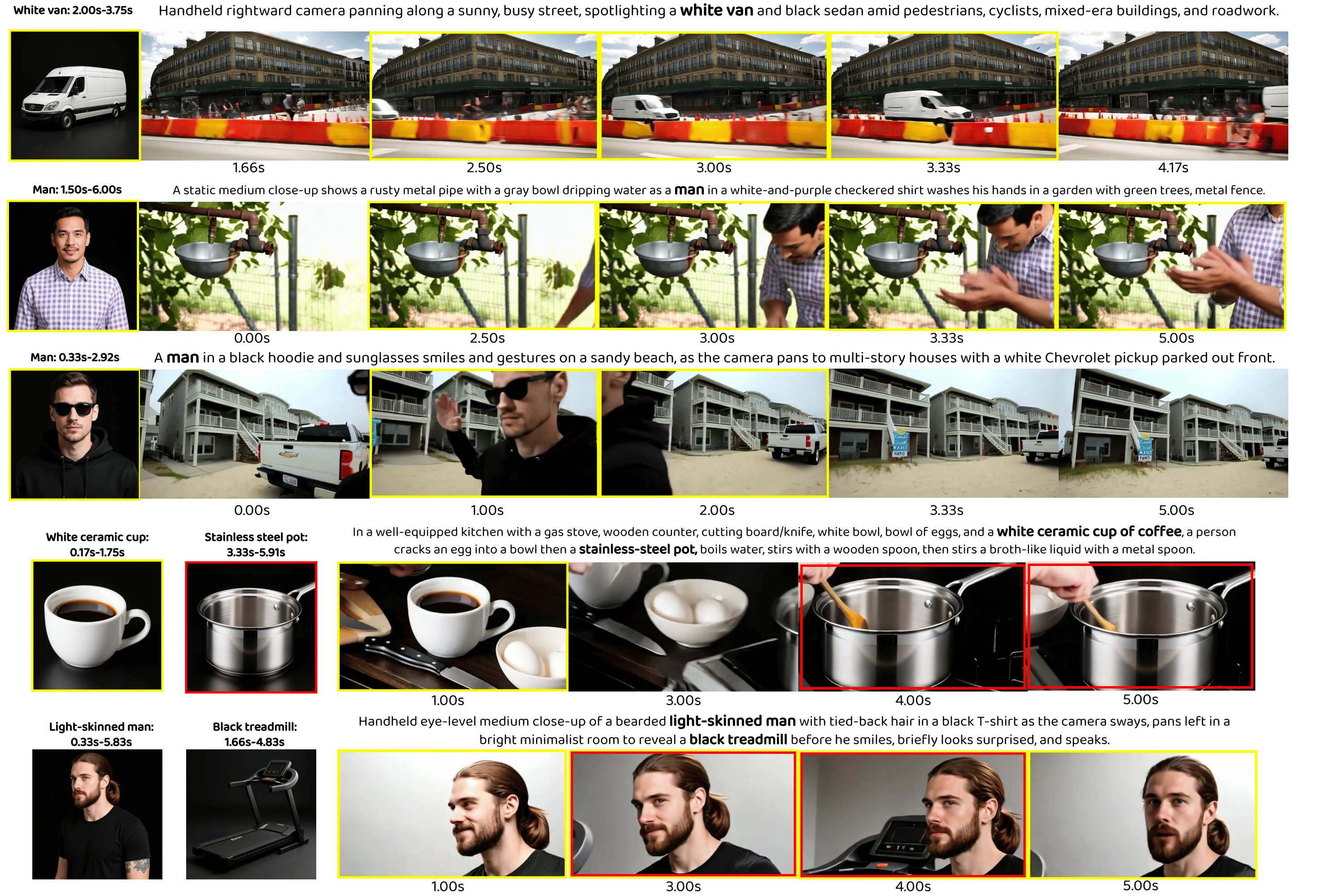}
  \captionof{figure}{\textbf{\ours: Time-controlled subject reference video generation.} Given a subject reference with input timestamps, \ours generates a consistent video with the subject naturally appearing in the specified time interval. Yellow boxes highlight the frames which lie in the input interval expecting the first subject reference to be present at those times, while red boxes highlight the second reference if present.}
  \label{fig:teaser}
\end{center}
}]

\begin{abstract}
Recent advances in subject-driven video generation with large diffusion models have enabled personalized content synthesis conditioned on user-provided subjects.
However, existing methods lack fine-grained temporal control over subject appearance and disappearance, which are essential for applications such as compositional video synthesis, storyboarding, and controllable animation.
We propose \ours, a unified framework that introduces explicit timestamps conditioning for subject-driven video generation.
Our approach introduces a novel positional encoding mechanism that unlocks the encoding of temporal intervals, associated in our case with subject identities, while seamlessly integrating with the pretrained video generation model positional embeddings.
Additionally, we incorporate subject-descriptive text tokens to strengthen binding between visual identity and video captions, mitigating ambiguity during generation.
Through token-wise concatenation, \ours avoids any additional cross-attention modules and incurs negligible parameter overhead.
We establish a benchmark evaluating multiple subject identity preservation, video fidelity, and temporal adherence.
Experimental results demonstrate that \ours achieves visual quality matching state-of-the-art video personalization methods, while, for the first time, enabling precise temporal control over multi-subject generation within videos. 
\end{abstract}

\section{Introduction}
\label{sec:intro}

Large-scale diffusion models~\cite{make_a_video,videoldm,magicvideo,videofusion,animatediff,cogvideox,snapvideo,sora} have shown remarkable quality and high-fidelity in producing realistic videos directly from text or image inputs. These models are capable of handling various forms of conditioning such as poses, depths, cameras~\cite{vace}.
More recently, conditioning based on identities or subjects has become popular to provide fine-grained control as well as personalized generations for users.
This has led to a large number of works targeting single or multiple reference conditions consisting of people, faces, animals, objects, or background~\cite{moonshot,videobooth,dreamvideo,customvideo,videodrafter,mcdiff,storydiffusion,materzynska2023customizing,customcrafter,id_animator,magicme,vimi,video_alchemist,magref,concept_master,bindweave,skyreels}.

However, existing works~\cite{video_alchemist,magref,concept_master,bindweave,skyreels} focus on conditioning for the full video duration.
In contrast, real-world videos consist of different events and shots composed together consisting of multiple subjects appearing at different points in the video.
This is particularly crucial for applications such as storyboarding, advertising, animations where different characters, logos appear at user-defined timestamps within a video.
Therefore, it is essential for a model to have fine-grained temporal control over when different subjects appear within the video.
Current subject reference models do not have direct control for timestep conditioning and will need to rely on textual prompts to do so.
However, current video models are incapable of understanding and strictly adhering to these types of prompts. 
Instead, the conditioning mechanism in reference models typically biases the generation toward videos in which the subjects appear continuously from the beginning to the end.

To this end, we propose \textbf{\ours}, a video generative model that accepts multiple subject references along with subject-specific temporal intervals, enabling precise control over each subject's appearance and disappearance within the generated video. 
Our model imposes no restrictions on subject type and can accommodate a wide range of open-set entities.
Prior works~\cite{video_alchemist,vimi,ip_adapter,tora2,magref,concept_master} have explored various combinations of self-attention and cross-attention to inject reference conditions using encoders such as CLIP~\cite{clip}, DINO~\cite{dino_v2}, face encoders~\cite{arcface}, or native VAEs.
In contrast, we find that directly encoding each reference input with the VAE and concatenating the resulting tokens with the video tokens provides an effective balance between quality and architectural simplicity.
This design removes the need for additional cross-attention parameters and instead fuses all inputs into a single unified token stream. 
Furthermore, because DiTs naturally support variable token counts, our model scales gracefully to an arbitrary number of input views. 
Finally, this results in a model with the feature spaces between references and video tokens which are well aligned leading to generations with high identity preservation.
Building on this foundation, we introduce an elegant mechanism for controlling temporal intervals of subject conditions.
This mechanism preserves the pretrained video model's original feature space while enabling precise control over when reference entities appear and disappear.
To encode the temporal boundaries of each subject, we modify the positional embedding mechanism for reference tokens by weighted blending of RoPE frequencies from the center and the two edges of the specified interval.
This design biases video tokens inside the interval to attend strongly to the corresponding reference, while attention strength decays smoothly outside the interval, producing natural transitions as subjects enter and exit the scene.

Finally, we introduce a large-scale data collection pipeline for obtaining high-quality data with precise timestamp labels for multiple subjects within a video.
Our pipeline consists of multiple stages of entity word detection, reference image detection at multiple video frames, followed by segmentation and tracking.
This results in each video pair consisting of multiple entities and masks tracking each entity over the duration of the video obtaining timestamp labels for the presence of different entities within a video.
During training, this allows us to selectively sample a subject frame which is far away from the sampled frames for the target video, directly resulting in complex augmentations for the subject with variations in pose, lighting, position, etc.
Coupled with other standard augmentations such as blur, zoom, rotation, flipping, repositioning~\cite{video_alchemist}, this prevents the model from copy-pasting the reference even when directly encoded with the same VAE.

Existing benchmarks for Subject-to-Video (S2V) generation, do not measure the timestamp following of subjects.
We introduce \textbf{\benchmark}, a benchmark for time-stamp conditioned Subject-to-Video generation.
This benchmark consists of an evaluation protocol for measuring subject identity preservation, text-video similarity, video fidelity, as well as timestamp following of subjects within the video.
Experiments comparing our approach to prior S2V works show that we perform favorably in terms of achieving personalized video generations with high fidelity, both quantitatively and qualitatively, while outperforming prior works in achieving timestamp control for S2V generation.

Our contributions are summarized as follows:
\begin{itemize}
\item We introduce \ours, a video generation model that supports multi-subject personalization with timestamp-based conditioning.
\item We devise a lightweight modification to reference-conditioned video generation models that enables control over an object's presence duration in a video, based on a weighted combination of RoPE frequencies.
\item We present a data collection pipeline for extracting subject timestamps, along with an evaluation benchmark to assess video quality, subject fidelity, and timestamp adherence.
\end{itemize}

\section{Related Works}
\label{sec:related_works}

\textbf{Personalization in Generative Models.}
One of the central goals of generative modeling research is achieving controllable generation by conditioning on modalities beyond text. Users often wish to see themselves or objects related to them represented in the generated content. The most natural conditioning signal for such applications is a reference image. Early works in this field employed optimization-based algorithms to encode reference images~\cite{dreambooth,textual_inversion,multi_concept_customization,han2023svdiff,consistory,avrahami2023break,jones2024customizing,layercomposer}. Later, encoder-based approaches were developed to eliminate or significantly reduce test-time fine-tuning~\cite{instantbooth,ip_adapter,arar2023domain,gal2023encoder,elite,li2023blip,chen2024anydoor,xiao2024fastcomposer,valevski2023face0,hyperdreambooth}.
Our approach falls into this second category, as it does not require test-time optimization. However, unlike these methods, our approach does not rely on a specialized encoder; instead, it utilizes the native latent diffusion VAE directly.

\myparagraph{Video Personalization.}
Several recent works extend image personalization methods to the video domain~\cite{moonshot,videobooth,dreamvideo,customvideo,videodrafter,mcdiff,storydiffusion,materzynska2023customizing,customcrafter,id_animator,magicme,vimi,video_alchemist,magref,concept_master,bindweave,skyreels}. Early methods primarily focused on limited domains, such as faces~\cite{id_animator,magicme} or single subjects from specific categories~\cite{moonshot,videobooth,dreamvideo,storydiffusion,customcrafter}. More recent studies explore an open-set, multi-subject paradigm, where multiple objects of arbitrary categories can be used for personalization.
This setting raises a key question: how to optimally inject identity while maintaining consistent text binding. One of the first works to explore this paradigm, \textit{Video Alchemist}~\cite{video_alchemist}, proposed a module that fuses each reference image with its corresponding subject-level text prompt. Later, \textit{SkyReelsA2}~\cite{skyreels} introduced a joint image–text embedding model for injecting multi-element representations into the generative process. \textit{Concept Master}~\cite{concept_master}, on the other hand, binds visual representations from CLIP~\cite{clip} with corresponding text embeddings for each concept through a Decoupled Attention Module (DAM). Concurrent work, \textit{Tora2}~\cite{tora2}, combines subject personalization with motion embeddings derived from trajectories using a gated self-attention mechanism. \textit{MAGREF}~\cite{magref} proposes a region-aware masking mechanism that merges references into a composite image encoded with a VAE.

While many prior works focus on optimizing identity injection and text–image binding mechanisms, we find that simple sequence-wise concatenation for identity injection, combined with a learnable embedding for text–image binding, achieves a sufficient level of fidelity. Therefore, our work shifts focus toward enriching video personalization through improved temporal control.

\myparagraph{Temporal Control.}
Compared to images, videos provide an additional temporal dimension that can be manipulated by generative models. Indeed, controlling event sequences and the timing of their occurrences is crucial for video generation. Recent studies propose various mechanisms for manipulating this temporal dimension. The pioneering work \textit{StoryBench}~\cite{storybench} introduced a benchmark for evaluating video generation models in sequential event generation, along with several autoregressive baselines. Subsequent works attempted to control event timing at inference time~\cite{oh2025mevg,kim2025tuningfreemultieventlongvideo}; however, such approaches often introduce a train–inference mismatch, leading to artifacted generations.
\textit{MiNT}~\cite{wu2025mint}, in contrast, trains the video model with an explicit time-control mechanism enabled by a novel cross-attention layer and a time-based positional encoding scheme (ReRoPE), which makes attention time-aware. Although effective, ReRoPE has a critical limitation: it requires rescaling video token RoPEs according to event interval length. Thus, it is only applicable to non-overlapping event sequences and is incompatible with MM-DiT-like~\cite{mmdit} (self-attention-only) conditioning mechanisms.

In our setting, reference conditionings may overlap, for example, multiple reference objects may need to appear together in a video. Furthermore, we aim to use sequence-wise concatenation (\ie, MM-DiT-like~\cite{mmdit} conditioning) for identity injection to maximally leverage video model priors. To achieve this, we develop a new temporal conditioning mechanism based on the summation of weighted RoPE frequencies from both interval midpoints and edges, which we call \textbf{Weighted RoPE}.

\begin{figure*}[t]
  \centering
  \includegraphics[width=\linewidth]{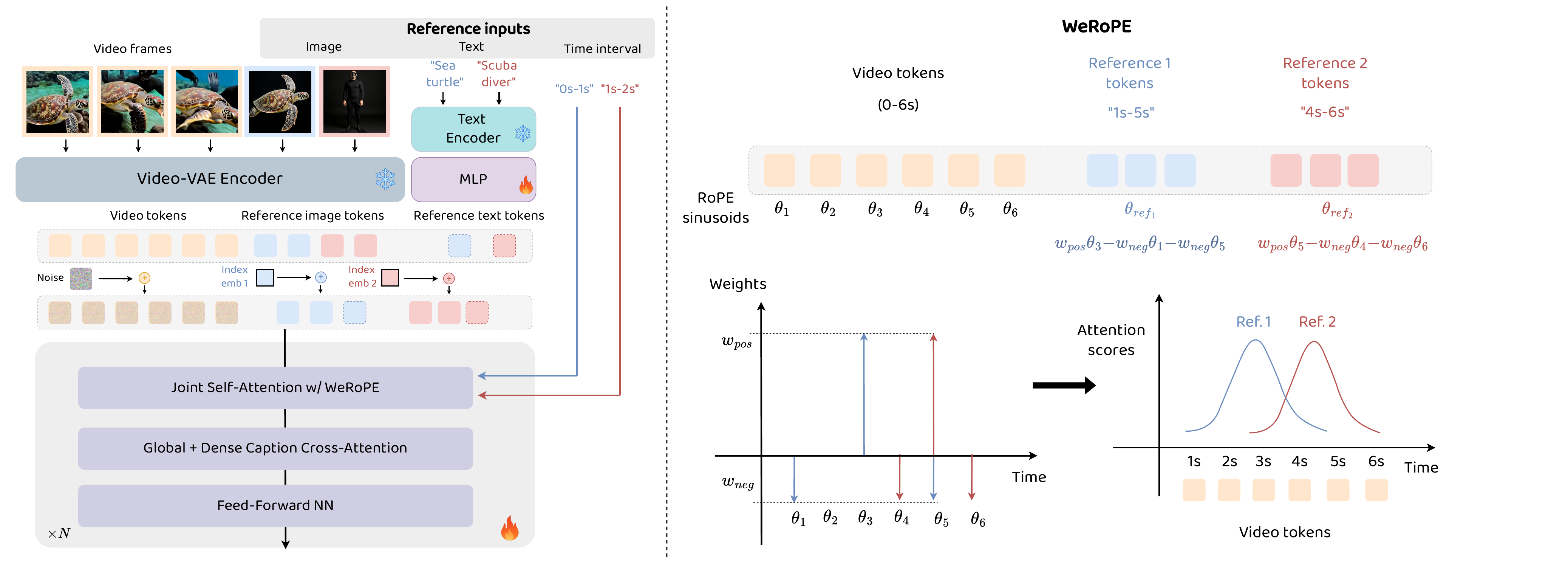}

  \caption{\textbf{Overview of our approach.} We design a video diffusion model to sequentially concatenate tokens from video, image and text references and jointly process them for multi reference-conditioned generation. Right: We develop a RoPE strategy for biasing attention scores of video tokens with reference tokens based on the input timestamps.}
  \label{fig:approach}
\end{figure*}

\section{Method}
\label{sec:method}

\subsection{Preliminary}

Our method builds on a pre-trained text-to-video latent diffusion backbone that combines a 3D variational autoencoder (VAE) with a Diffusion Transformer (DiT). A video
\(
\mathbf{x}\in\mathbb{R}^{T\times 3\times H\times W}
\)
is mapped to a latent tensor
\(
\mathbf{z}=\mathcal{E}(\mathbf{x})\in\mathbb{R}^{T'\times C\times H'\times W'}
\)
by the VAE encoder \(\mathcal{E}\), and decoded by \(\mathcal{D}\) during synthesis.

We adopt the rectified-flow formulation in latent space. Let \(p_0\) denote the data distribution of \(\mathbf{z}\) and \(p_1=\mathcal{N}(\mathbf{0},\mathbf{I})\) the standard Gaussian. For a diffusion time \(t\sim\mathcal{U}(0,1)\) and noise \(\boldsymbol{\epsilon}\sim p_1\), we form linear interpolants:
\begin{equation}
\mathbf{z}_t=(1-t)\,\mathbf{z}+t\,\boldsymbol{\epsilon}.
\end{equation}

The DiT predicts the target velocity field
\(
\mathbf{v}^\star=\boldsymbol{\epsilon}-\mathbf{z}
\)
conditioned on the timestep and text embeddings \(c_{\text{text}}\). The training objective is the flow-matching loss
\begin{equation}
\begin{aligned}
\mathcal{L}_{\text{flow}}
= \mathbb{E}_{%
  \substack{
    \mathbf{z}\sim p_0, \boldsymbol{\epsilon}\sim p_1,\\
    t\sim\mathcal{U}(0,1)
  }}
\left\|
\mathbf{v}_\theta(\mathbf{z}_t, t, c_{\text{text}})
- (\boldsymbol{\epsilon}-\mathbf{z})
\right\|_2^2
\end{aligned}
\end{equation}
where \(\mathbf{v}_\theta\) is the DiT parameterized by \(\theta\). Classifier-free guidance is applied in the usual way by randomly dropping \(c_{\text{text}}\) during training.

The DiT backbone consists of a patchifier with spatial downsampling followed by a stack of spatiotemporal self-attention blocks interleaved with text cross-attention and feed-forward layers. Temporal and spatial positions are encoded via Rotary Positional Embeddings (RoPE).

\subsection{Model Architecture}
In this section, we outline the architectural details of our model. \ours is a DiT-based framework which takes in image inputs, timestamp intervals and entity words describing the image (\textit{e.g.} dog, man, pen). Therefore, the model takes in N inputs $[(\boldsymbol{I}_n,[t_0^n,t_1^n],w_n)]_{n=1}^N$. The goal of our model is to generate a video which is consistent with the input references appearing between their specified timestamps while also maintaining high video fidelity.

Many prior works~\cite{video_alchemist, concept_master, tora2, ip_adapter} with subject conditioning propose an additional Cross-Attention layer for DiTs which encode image references with semantic encoders such as DINO~\cite{dino_v2}, CLIP~\cite{clip}, ArcFace~\cite{arcface}, and so on. While such encoders enable the model to utilize semantic information from the image, they require fairly complex design for fusing image and text reference information into the DiT via IP-Adapters, Q-formers, additional Attention blocks, resulting in additional parameters to the model. We instead propose to do joint processing of video tokens and the subject reference. We directly encode the image references with the VAE and sequentially concatenate the reference tokens with the video tokens. This allows us to simultaneously process both streams of tokens, without any additional parameter cost, and ensuring the feature spaces between video and image reference remain aligned. This helps maintain stronger identities compared to the cross-attention counterparts which may miss encoding certain attributes from the image based on the type of encoder used. Finally, another key benefit is enabling temporal control of reference tokens by varying the RoPE frequencies for the reference tokens.

Video DiTs use 3D Rotary Positional Embeddings (RoPE)~\cite{rope} for allowing the transformer architecture to understand the position \((x,y,t)\) of video tokens.
For each latent token \(z_{xyt}\), RoPE applies a frequency-based rotation
\begin{equation}
\hat{z}_{xyt} = \textnormal{RoPE}(z, x, y, t) = R\!\left(z_{xyt},\,\theta_{xyt}\right),
\end{equation}
where \(\theta_{xyt}\) denotes the sinusoidal phase parameters (see supplementary for details). With the relative nature of RoPE, it naturally decays the attention scores between tokens reducing their dependency with each other as the distance between them increases. This provides a natural mechanism for controlling the attention scores between reference tokens and video tokens. We aim to develop a RoPE mechanism for reference tokens which maintains high attention scores with the video tokens within the interval but naturally falls off outside of it.

\begin{figure}[t]
  \centering
  \includegraphics[width=\linewidth]{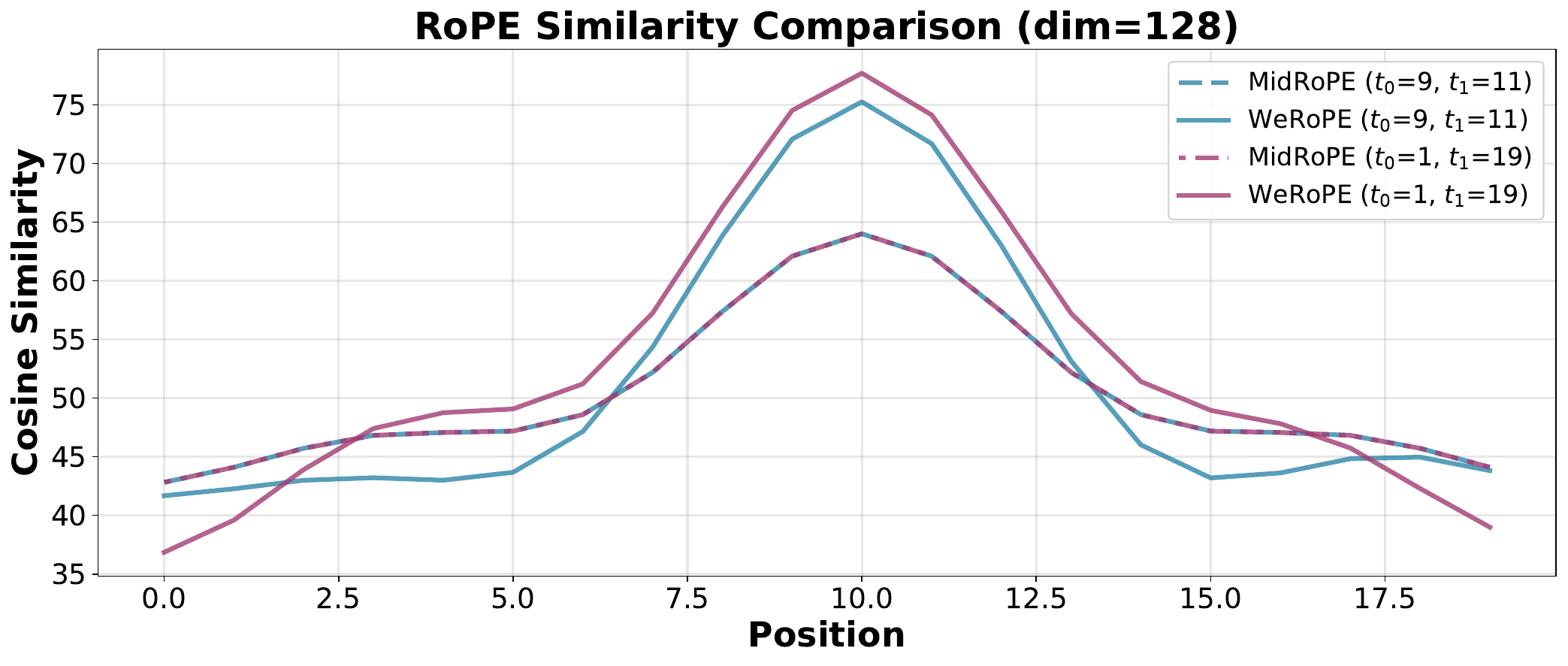}
  \caption{\textbf{Effect of MidRoPE \textit{v.s.} WeRoPE.} }
  \label{fig:rope_comparison_decay}
  \vspace{-6pt}
\end{figure}

For ease of notation, without any loss of generality, we restrict ourselves to a single reference image consisting of tokens $\boldsymbol{r}$. Each reference image has the same number of tokens as a single latent frame $\boldsymbol{z}_t$ in the video. To maintain spatial relationships between tokens within the reference and to enable variable temporal attention scores, we maintain 2D spatial RoPE but modify the time frequency for the reference tokens. One solution is to utilize the frequency corresponding to the midpoint of the time interval $t_{m}=(t_0+t_1)/2$. This yields us,
\begin{equation}
    \hat{r}_{xy} = \textnormal{RoPE}(r, x, y, t_{mid})
\end{equation}
We term this type of rope as MidRoPE. We visualize the attention decay for the reference RoPE with respect to video RoPE for two separate time intervals in~\cref{fig:rope_comparison_decay}. While MidRoPE centers the reference tokens at the midpoint of the interval with a decay, it does not control the rate of decay leading to ambiguity for intervals with the same midpoint (\eg, $[8,10]$ or $[1,17]$). We instead propose a weighing strategy combining the frequency at the midpoint with the frequencies at the edge of the interval. 

\myparagraph{Weighted RoPE (WeRoPE).} We define a set of weight hyperparameters $w_p,w_n$ where $w_p$ sets the weight of the frequency at $t_{mid}$ while $w_n$ sets the weight of the frequency at its edges. We define $t_l=t_0/2, t_r=(T-t_1)/2$ as the timestamps at the left and right side of interval, where T is the number of latent frames. 
Due to the linearity of RoPE, we then obtain
\begin{equation}
\begin{aligned}
\hat{r}_{xy}
&= R\big(r_{xy},\, w_p\theta_{xyt_{mid}} + w_n(\theta_{xyt_l}+\theta_{xyt_r})\big) \\
&= w_p R(r_{xy},\theta_{xyt_{mid}})
  + w_n\!\!\sum_{t_i\in\{t_l,t_r\}}\!R(r_{xy},\theta_{xyt_i})
\end{aligned}
\end{equation}
We then set a positive value for $w_p$ and negative value for $w_n$ which leads to a weighted combination of multiple RoPE decay profiles into one as shown in~\cref{fig:rope_comparison_decay}. As the intervals change for each reference, we can bias the attention scores directly via WeRoPE without any other form of conditioning.

\myparagraph{Multi-reference Disentanglement and Textual Binding.} Multiple reference inputs for the model introduce ambiguity in the generations especially when a caption contains multiple reference entities of the same class (\eg, man or woman). Furthermore, the network does not include any positional information between references. To combat this, we first incorporate learnable embeddings for each index of the reference which disentangle two tokens at the same spatial location from different references. Furthermore, we include reference word tags which uniquely identify each reference and encode it with the text encoder used for the base caption as additional input tokens to the DiT. We apply a small MLP to the text embeddings as they lie in a different feature space compared to the video or reference latents. We apply the same index embedding to the word tag for each reference in order to bind the word and the corresponding image. We maintain the same temporal WeROPE for the text and apply Spatial RoPE in a diagonal fashion as is done in MMDiTs like Qwen-Image~\cite{qwen_image_edit}. 
As the word tag embedding undergoes further processing and Cross-Attention with the video caption embeddings, it allows for the network to bind the words in the captions with the word tags and the reference images. \cref{ssec:ablations} shows that this word tag binding is essential for decoupling similar identities.

\subsection{Data Collection Pipeline}
\label{ssec:data_pipeline}

Training the model for time-stamp conditioned references requires a dataset with videos, reference identities and timestamps. Due to the lack of such high-quality data, we propose a new data collection pipeline with multiple stages briefly discussed below.  We provide further details in the supplementary.

We start with a large corpus of text and video pairs. Our dataset consists of global captions describing the full scene and timed local/dense captions similar to the dataset used in~\cite{wu2025mint}.
To obtain multiple entities, we query an LLM~\cite{bai2023qwenvlversatilevisionlanguagemodel} to retrieve word tags referencing different entities within a video caption. We include several criteria for removing keywords such as body parts, groups of entities, large-scale scenes which cannot be segmented, and so on. We additionally force word tags to be unique to remove ambiguities.

For each entity, we use Grounding Dino to detect bounding boxes at several timestamps within the video. This allows us to obtain a high recall in detecting each entity within the video. We keep the bounding box with the highest CLIP similarity score with the entity word tag. We subsequently run SAM2 to track the entity across the video in forward and reverse directions. This gives us a set of mask tracks for each entity.

During training, we compute a time-stamp interval for a reference by identifying the first and last frame of the mask occurring in the video over a certain threshold of pixels. Furthermore, by storing framewise masks for each reference, we sample the masks outside the timestamps sampled for the video at any given iteration. This acts as a strong augmentation in choosing references with different poses/lighting compared to the ones in the video frames. We additionally include other augmentations such as blur, zoom, color jitter, and so on as proposed in~\cite{video_alchemist}. We additionally crop the image for the sampled mask and center it which prevents the model from copy-pasting via the spatial location bias from RoPE.

\section{Experiments}
\label{sec:experiments}
We describe the experimental details in~\cref{ssec:implementation_detail}, discuss quantitative and qualitative evaluations of our benchmark in comparison to prior work in~\cref{ssec:results}, and ablate the different components of our model in~\cref{ssec:ablations}.

\subsection{Implementation Details}
\label{ssec:implementation_detail}

\begin{figure*}[t]
  \centering
  \includegraphics[width=0.98\linewidth]{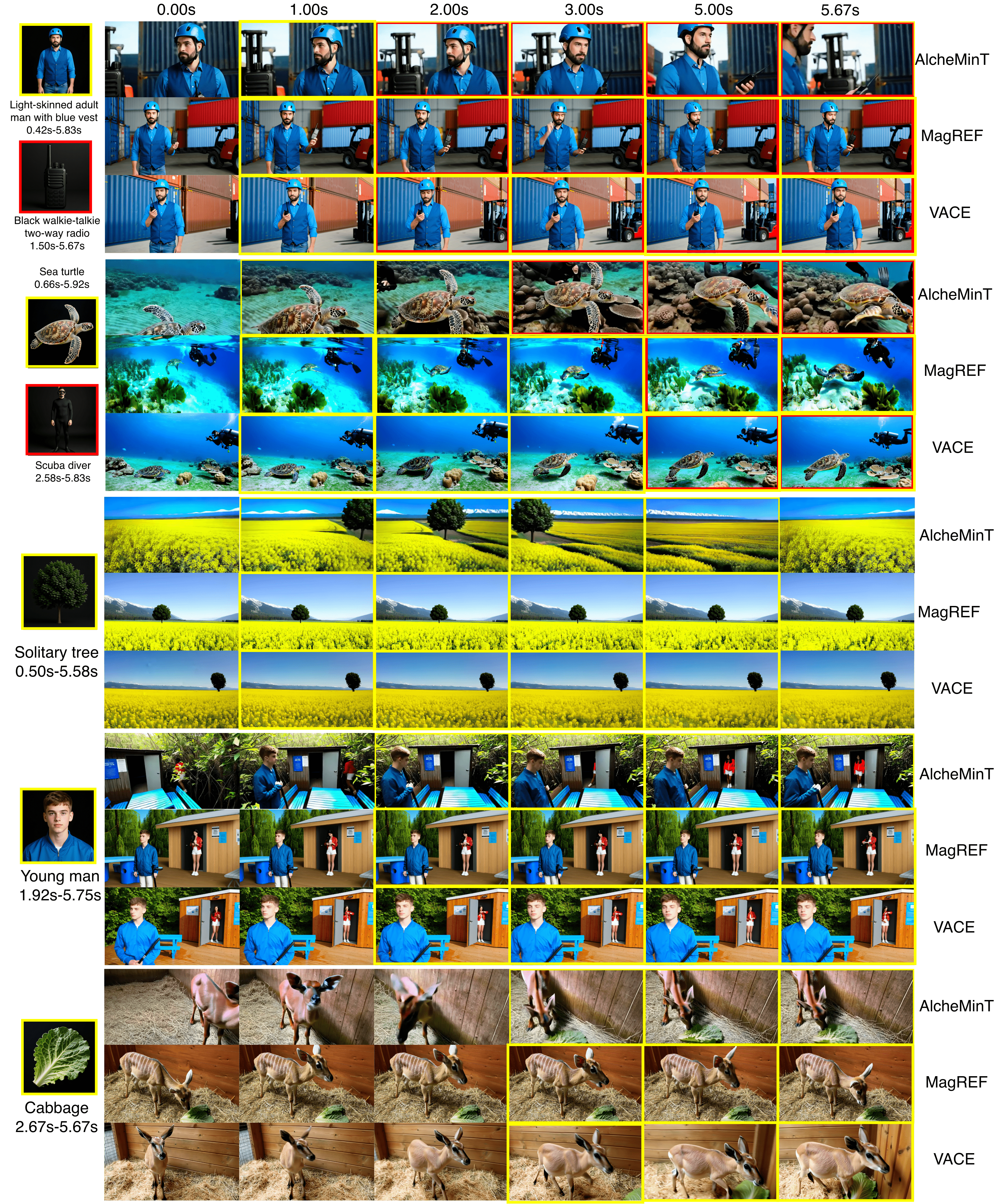}
  \caption{\textbf{Qualitative visualizations.} Yellow boxes indicate the expected occurence for the first reference and red boxes for the second. We obtain high fidelity generations which closely follow the input time interval and preserving the subject identity while prior works struggle to control the appearance/disappearance of these references.}
  \label{fig:qualitative}
  \vspace{-6pt}
\end{figure*}

\myparagraph{Model Training.}
We train our model using the dataset collected from the pipeline described in~\cref{ssec:data_pipeline}. We fully finetune our model on a base text-to-video DiT model. We incorporate additional parameters with subject index embedding, text embedding MLP, and a parallel cross attention branch for dense captions with ReRoPE~\cite{wu2025mint}.  These parameters are finetuned along with the base model parameters with a learning rate of $1.0e^{-4}, 3.0e^{-5}$ respectively linearly warmed up for 1K iterations. Our model is trained for 30K iterations on 16 80GB H100 GPUs with a batch size of 32. We randomly drop reference conditions, jointly for both image and text, for enabling CFG along with video caption. We do not zero out corresponding timestamp intervals when dropping reference intervals as we find altering WeRoPE for the unconditional pass introduces artifacts during generation. Inference is done with rectified flow sampling for 40 steps with a time-shifting value of 5.66. We use different CFG values for reference, text, and both, described in more detail in the supplementary.

\myparagraph{Evaluation and Benchmark}. While several existing benchmarks exist~\cite{video_alchemist,yuan2025opens2v,concept_master} for measuring quality of subject references in video generation, they do not measure the time-stamp interval for the reference in the generated video. We therefore propose a new benchmark dubbed \benchmark, incorporating existing metrics measuring subject identity preservation while also their generated time-stamp interval. The benchmark dataset consists of a set of textual prompts which are fed to an LLM model with an instruction to output upto 2 reference entities from the text while also following plausible timestamps.  We additionally also generate global and dense captions from the base prompt and a description of the entity which is consistent with the base prompt. We then provide this entity description as part of a prompt template to a Text-to-Image model~\cite{qwen_image_edit} producing reference images. Unlike prior benchmarks with fixed entity keywords consisting of common objects (such as cup, ball, man), our dataset can consist of open-set entities which can describe subjects, objects, scenes, and so on. Our evaluation protocol utilizes Grounding Dino and SAM2 to track entities through the video. We obtain bounding boxes from multiple frames within the interval, keeping the one with the highest CLIP score which is subsequently fed to SAM2 to produce a full mask track. We obtain time intervals for masks above a predefined area threshold as the predicted intervals which we compare with the GT interval. We measure IoU overlap (t-IOU) as well as the L2 error (t-L2) between the start and end indices of predicted and GT intervals, normalized between 0 and 1. To measure identity preservation, we extract average CLIP scores between the segmented masks and the reference text (\clipt) as well as the CLIP scores between the segmented masks and the reference image (\clipr). For a fair comparison with the baselines, we augment the prompt to the model as ``${<}$entity word${>}$ appears between ${<}t_0{>}$ seconds and ${<}t_1{>}$ seconds'' as a straightforward way of encoding time intervals into the generation.

\subsection{Results}
\label{ssec:results}
\begin{table}
    \caption{\textbf{Quantitative evaluation on \benchmark.} We compare against SOTA multi-reference video generation approaches achieving best time-following capability.}
    \label{tab:quantitative_ours}
    \small
    \setlength{\tabcolsep}{2.5pt}
    \begin{tabular}{l | c | c c c c}
        \toprule Approach & \# Refs & t-L2$\downarrow$ & t-IOU$\uparrow$ & \clipt $\uparrow$& \clipr$\uparrow$\\ %
        \midrule

        MAGREF~\cite{magref} & \multirow{3}{*}{1}&0.332 & 0.404 & \textbf{0.257} & 0.791 \\
        VACE~\cite{vace} && 0.340 & 0.396 & 0.258 & \textbf{0.797} \\
        SkyReels~\cite{skyreels} && 0.318 & 0.421 & 0.262 & 0.775 \\
        \ours (Ours) && \textbf{0.281} & \textbf{0.433} & 0.252 & 0.768 \\
        \midrule
        MAGREF~\cite{magref} & \multirow{3}{*}{2}&0.365 & 0.353 & 0.260 & \textbf{0.803} \\
        VACE~\cite{vace} && 0.354 & 0.368 & 0.258 & 0.798 \\
        SkyReels~\cite{skyreels} && 0.341 & 0.380 & \textbf{0.261} & 0.782 \\
        \ours (Ours) && \textbf{0.291} & \textbf{0.413} & 0.253 & 0.775 \\
        \bottomrule
    \end{tabular}
\end{table}

\myparagraph{Quantitative Evaluation.}
We evaluate our approach quantitatively on our benchmark against the SOTA Multi-Subject to Video methods of~\cite{magref,vace}. Results are summarized in~\cref{tab:quantitative_ours}. We obtain consistently better results in terms of timestamp metrics of t-L2 and t-IOU for both single and multi-reference case. We obtain comparable performance in terms of \clipt but slightly worse compared to~\cite{magref} with lower reference fidelity. This, however, stems from the base video model with higher capacity leading to finer detail in the subject. Nevertheless, this highlights the capability of the model to produce high fidelity video generations with the subject reference consistent with the input.

\myparagraph{Qualitative Evaluation.}
We also present qualitative visualizations for the benchmark in~\cref{fig:qualitative}. The improvement in t-L2, t-IoU metrics are supported by the visual results where our videos consistently follow the input interval with little error across different samples. Notably, we are able to generate videos consistent with the input time intervals with a combination of subject motions (such as the walkie-talkie, scuba diver), camera motions (such as the tree, cabbage, or the young man), or both (sea turtle). We see that we are able to obtain smooth videos without abrupt motion and with high-fidelity subject consistent generations. In contrast, prior works of~\cite{magref,skyreels} fail to follow the input timestamp condition for different references. This leads to relatively more static videos with subjects present throughout the video.

\subsection{Ablations}
\label{ssec:ablations}
We now ablate different components of our approach. To isolate the effect of each component, we train model variants with the global and dense captions concatenated together without any dense cross attention branch.

\begin{figure}[t]
  \centering
  \includegraphics[width=\linewidth]{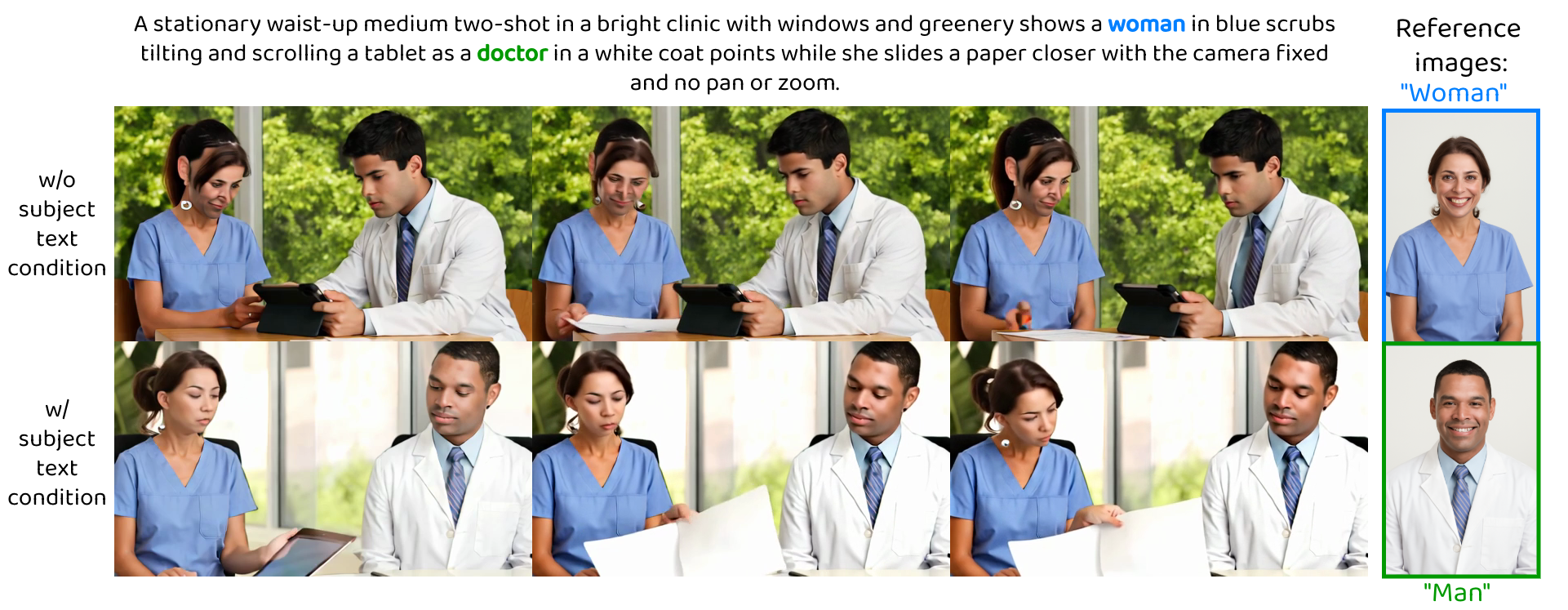}
  \caption{\textbf{Effect of reference text binding with image.} Including reference text embeddings allows the model to attend to the caption and bind the subject such as ``doctor'' in caption and ``man'' in reference text to produce aligned subjects in the generated video.}
  \label{fig:ablations_text_cond}
  \vspace{-6pt}
\end{figure}

\myparagraph{Reference Text Binding.} We train two variants of our model with and without reference text embeddings as input to the model. To measure the effect of text for multiple references we evaluate on a subset of our benchmark with 2 references. Results are summarized in~\cref{tab:ablation}. We see that including text embeddings improves the \clipr score as expected but lowers the \clipt score. This occurs because the generated reference masks remain broadly aligned with the reference text but lose fine-grained correspondence to the reference image. We visualize this in~\cref{fig:ablations_text_cond}, where the text-conditioned model maintains disentangled entities remain consistent with the input reference (\eg, associating “man” with “doctor”), while removing text conditioning leads to confusion between entities, visible as artifacts on the woman’s face and inconsistencies in the doctor’s appearance with respect to the reference image.

\begin{figure}[t]
  \centering
  \includegraphics[width=\linewidth]{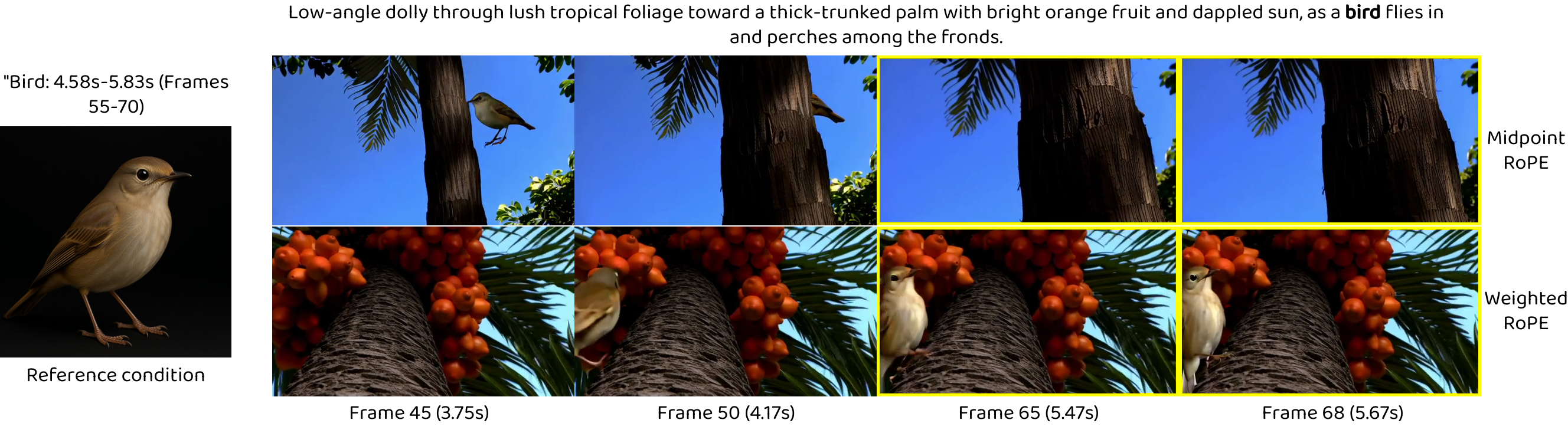}
  \caption{\textbf{Comparison of MidRoPE and WeRoPE.} WeRoPE allows the model to incorporate the reference condition (bird) adhering to the input interval while Midpoint RoPE is incapable of doing so without interval length information. Yellow boxes are highlighted when frames lie in the input interval.}
  \label{fig:ablations_werope}
  \vspace{-6pt}
\end{figure}

\myparagraph{Comparison of MidRoPE and WeRoPE.} We now ablate the effect of the two RoPE variants on our benchmark. We train the 2 models with a single reference as condition while also evaluating on this setting in our benchmark. We see that WeRoPE improves upon both timestamp error metrics as expected but results in lower CLIP scores. WeRoPE provides missing information about the reference interval through the negative sampled indices yielding better time-stamp following via lower t-L2 and higher t-IOU while also better preserving the identity information. We visualize this in~\cref{fig:ablations_werope} for a short timestamp interval input between 4.58s and 5.83s for the reference image ``Bird''. MidRoPE generates the bird at the start of the video, disappearing during the specified timestamps. WeRoPE  shows the bird flying in closer to the start of the input interval at time 4.58s.

\begin{table}
    \caption{\textbf{Ablation study.} We ablate the effect of incorporating text conditioning into the network as well as the 2 RoPE variants of MidRoPE and WeRoPE}
    \label{tab:ablation} \small
    \renewcommand{\arraystretch}{1.0}
    \setlength{\tabcolsep}{3pt}
    \begin{tabular}{l | c c c c}
        \toprule Variant & t-L2$\downarrow$ & t-IOU$\uparrow$ & \clipt$\uparrow$ & \clipr$\uparrow$\\ %
        \midrule
        w/o Ref. Text Embedding& 0.139 & 0.751 & \textbf{0.216} & 0.718 \\
        w/ Ref. Text Embedding & \textbf{0.135} & \textbf{0.755} & 0.214 & \textbf{0.724} \\ %
        \midrule
        MidRoPE & 0.300 & 0.453 & \textbf{0.227} & \textbf{0.715} \\
        WeRoPE & \textbf{0.288} & \textbf{0.469} & 0.216 & 0.691 \\ %
        \bottomrule
    \end{tabular}
\end{table}

\section{Conclusion}
In this work, we proposed \ours, a new framework for multi-reference conditioned video generation with additional fine-grained temporal inputs for controlling the appearance of these subjects. We develop a novel architecture for sequentially concatenating reference latents with video latents encoded with the same VAE along with reference text embedding latents, binding the video captions and subject, and reducing generation ambiguities.

Furthermore, we introduce a novel RoPE mechanism for transforming the frequencies for the references based on the input time intervals, thereby temporally biasing the attention scores with the video tokens. We propose a data collection pipeline for obtaining subject timestamp intervals as well as develop a benchmark for evaluating on this new task. Through qualitative and quantitative results on our benchmark, we show that, compared to prior works, we obtain much better timestamp following of the subjects while maintaining high subject identity and video fidelity. This opens the avenue for long-video generation with time-controlled references which is important for various industrial applications such as advertisements, storyboarding, and so on.

\appendix
\clearpage
\setcounter{page}{1}
\maketitlesupplementary

~\cref{supp_sec:baselines} provides additional baseline comparisons and with longer training of our approach.~\cref{supp_sec:ablations} includes further ablations with variants more close to our final approach.~\cref{supp_sec:data_collection} discusses the details of our data collection pipeline in depth,~\cref{supp_sec:multi_cfg} provides information about our inference CFG setup with multi text and reference conditions, and~\cref{supp_sec:rope} provides background information for RoPE. We provide qualitative visualizations in~\cref{supp_sec:addn_qualitative}. Please refer to videos (or anonymous HTML file) attached in the supplementary for additional generations from our approach.

\section{Additional baselines and longer training}
\label{supp_sec:baselines}
In this section, we compare against existing baselines~\cite{magref,vace,skyreels} in Table 1 of the main paper as well as~\cite{video_alchemist}. We additionally also perform training for an additional 15K iterations and evaluate on a variation of the benchmark with captions allowing for multiple references entering/leaving the scene with lengths between 0.5 and 4.5 seconds.
Results are summarized in~\cref{supp_tab:quantitative_ours}. With longer training, we achieve even better temporal following while also outperforming prior works for subject identity preservation.

\begin{table}
    \caption{\textbf{Quantitative evaluation on \benchmark.} We compare against SOTA multi-reference video generation approaches achieving best time-following capability while matching identity preservation performance.}
    \label{supp_tab:quantitative_ours}
    \small
    \setlength{\tabcolsep}{2.5pt}
    \begin{tabular}{l | c | c c c c}
        \toprule Approach & \# Refs & t-L2$\downarrow$ & t-IOU$\uparrow$ & \clipt $\uparrow$& \clipr$\uparrow$\\ %
        \midrule

        MAGREF~\cite{magref} & \multirow{4}{*}{1}&0.284 & 0.518& 0.231& 0.724 \\
        VACE~\cite{vace} && 0.257 & 0.537 & 0.230 & 0.740 \\
        SkyReels~\cite{skyreels} && 0.278 & 0.497 & 0.232 & 0.733 \\
        Alchemist~\cite{video_alchemist} && 0.300 & 0.469 & 0.250 & 0.770 \\
        \ours (Ours) && \textbf{0.217} & \textbf{0.568} & \textbf{0.258} & \textbf{0.784} \\ %
        \midrule
        MAGREF~\cite{magref} & \multirow{4}{*}{2}&0.283 & 0.525 & 0.230 & 0.730 \\
        VACE~\cite{vace} && 0.263 & 0.525 & 0.230 & 0.730 \\
        SkyReels~\cite{skyreels} && 0.268 & 0.512 & 0.232 & 0.718 \\
        Alchemist~\cite{video_alchemist} && 0.298 & 0.477 & 0.250 & 0.764 \\
        \ours (Ours) && \textbf{0.235} & \textbf{0.552} & \textbf{0.260} & \textbf{0.776} \\
        \bottomrule
    \end{tabular}
\end{table}

\section{Ablations}
\label{supp_sec:ablations}
We ablate the design of our pipeline, namely text conditioning, as well as WeRoPE, on our benchmark similar to Table 2 in the main paper but also including dense captions via CrossAttention + ReRoPE~\cite{wu2025mint}. For a fair comparison, we compare with models trained for 25K iterations on a subset of the test set with 2 references, to highlight the effect of disentanglement. We include the baseline setting of no temporal RoPE for the reference tokens as well, effectively having a $0$ timestamp. Results are summarized in~\cref{supp_tab:ablation}. Similar to the observations in Sec. 4.3 of the main paper, we see that including reference text embeddings improves overall subject preservation with higher \clipt, \clipr while maintaining similar timestamp following with lower t-L2, but also marginally lower t-IOU. This is due to text conditioning not affecting timestamp following of the generated video but to disambiguate appearances between multiple subjects as highlighted by the CLIP scores. Additionally, not providing any temporal RoPE for reference tokens increases the error in temporal following even further highlighting the importance of the RoPE mechanism for controlling the timing of the references.

\begin{table}
    \caption{\textbf{Ablation study.} We ablate the effect of incorporating text conditioning into the network as well as the 2 RoPE variants of MidRoPE and WeRoPE, by utilizing dense captions~\cite{wu2025mint} into the model.}
    \label{supp_tab:ablation} \small
    \renewcommand{\arraystretch}{1.0}
    \setlength{\tabcolsep}{3pt}
    \begin{tabular}{l | c c c c}
        \toprule Variant & t-L2$\downarrow$ & t-IOU$\uparrow$ & \clipt$\uparrow$ & \clipr$\uparrow$\\ %
        \midrule
        w/o Ref. Text Embedding& 0.313 & \textbf{0.419} & 0.231 & 0.724 \\
        w/ Ref. Text Embedding & \textbf{0.283} & 0.402 & \textbf{0.251} & \textbf{0.761} \\ %
        \midrule
        No RoPE & 0.360 & 0.323 & 0.234 & 0.728 \\
        MidRoPE & 0.336 & 0.346 & 0.221 & 0.702 \\
        WeRoPE & \textbf{0.294} & \textbf{0.368} & \textbf{0.250} & \textbf{0.757} \\ %
        \bottomrule 
    \end{tabular}
\end{table}

\begin{figure*}[t]
  \centering
  \includegraphics[width=0.95\linewidth]{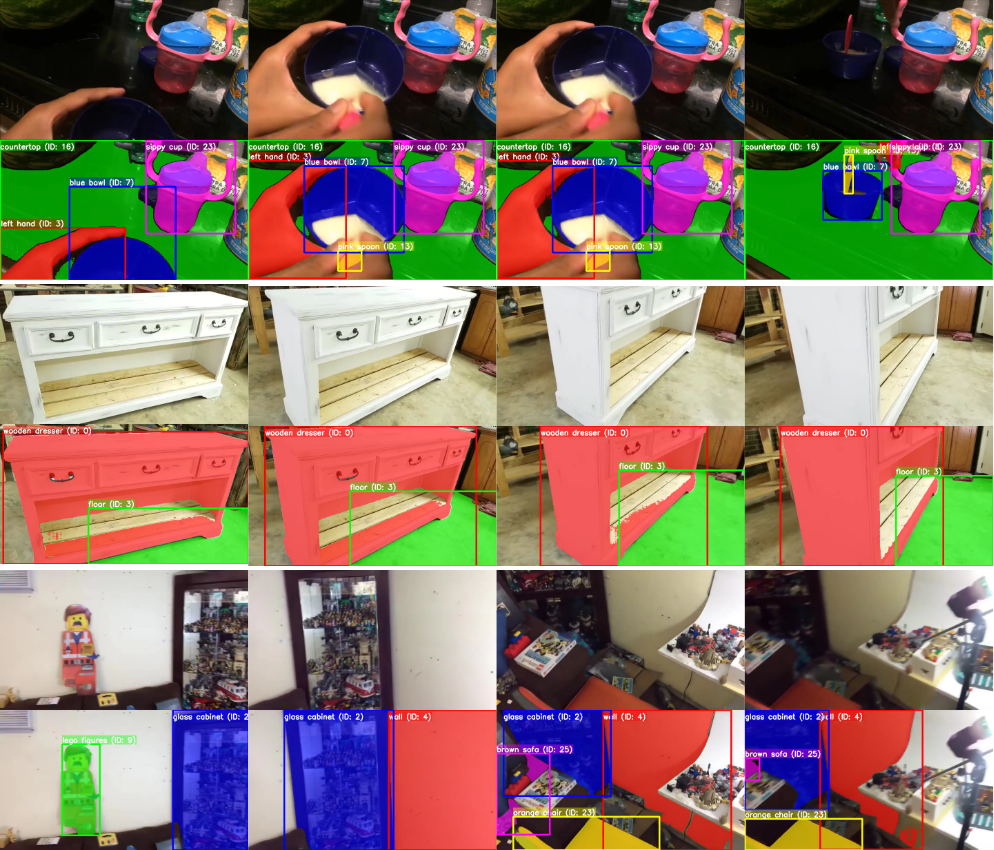}
  \caption{\textbf{Segmented track visualization.} We show segmented tracks obtained via our data collection pipeline for a wide variety of objects along with the original videos. Objects appearing/disappearing are tracked consistently across frames with complex word descriptions.}
  \label{supp_fig:data_viz}
  \vspace{-6pt}
\end{figure*}

\section{Data collection pipeline details}
\label{supp_sec:data_collection}
Our dataset is built on top of an existing video-text paired datasets. We additionally collect dense timed captions for each temporal event similar to~\cite{wu2025mint}. 

\subsection{Entity Word Extraction}
\label{supp_ssec:entity_word}
We extract, for every dense caption, the set of phrases that denote \emph{physically groundable} entities, similar to~\cite{video_alchemist}. We use Qwen~2.5~\cite{bai2023qwenvlversatilevisionlanguagemodel} with a fixed prompt template that includes the input caption and task description, and we augment the prompt with the following constraints:
\begin{itemize}
  \item Each entity phrase is an exact substring of the caption. 
  \item When multiple entities share similar labels, disambiguate using adjectives or referring expressions present in the caption.
  \item Exclude terms from a predefined blacklist (e.g., \texttt{letters}, \texttt{text}, and generic body parts such as \texttt{arm}, \texttt{leg}).
\end{itemize}

\noindent This procedure yields an open-vocabulary inventory of entity phrases rather than a closed set of classes, and it naturally supports reference disambiguation (e.g., the two men are distinguished by their local modifiers and roles in the caption). This also allows the model to bind word tags with the captions through our text conditioning strategy.

In practice, we occasionally observe intangible or non-physical descriptors (e.g., \textit{left}, \textit{right}, \textit{area}) or blacklist violations, likely due to the model jointly performing extraction and rule following. To address this, we apply a lightweight post-filtering pass: the extracted list is fed back to Qwen with instructions to remove blacklist items and any terms that do not denote physically groundable entities.

\noindent We provide an example input caption and output pair below.
\vspace{5pt}
\begin{lstlisting}[style=snippet]
Caption: The camera dollies back and pans to the left showing a full shot of a fair-skinned man with a short brown beard, wearing a red t-shirt, blue jeans with white shoes, handing over the keys to the fair-skinned man standing on the left. The man has short black hair and beard wearing a white t-shirt with grey jeans and grey shoes.
Output: ['fair-skinned man with a short brown beard', 'fair-skinned man standing on the left', 'red t-shirt', 'blue jeans', 'white shoes', 'short black hair', 'beard', 'white t-shirt', 'grey jeans', 'grey shoes']
\end{lstlisting}

\subsection{Grounded Entity Mask Tracking}
\label{supp_ssec:gdino_sam2}

To obtain temporally consistent instance masks for all entities across a video, we combine Grounding DINO~\cite{grounding_dino} for text-conditioned detections with SAM2~\cite{sam} for mask tracking.

\paragraph{Entity detections.}
For each entity keyword, we take its associated dense caption and time interval in the video. We sample the $10^{\textnormal{th}}, 50^{\textnormal{th}}, 90^{\textnormal{th}}$ percentile frames within that interval and run Grounding DINO using the keyword as the text prompt. Within each frame, we apply non-maximum suppression (NMS) to remove duplicate boxes, then select the remaining detection with the highest CLIP similarity to the keyword. Each selected detection stores the tuple: \{word tag, caption and interval, frame index, bounding box\}.

\paragraph{Tracking and mask propagation.}
For every detection, we invoke SAM2 with the detection box as a box prompt at the detection frame index and track the instance forward and backward to produce a per-frame mask track over the full interval.

\paragraph{Deduplication.}
Because multiple detections (from different dense captions) may correspond to the same physical entity, we compute the average mask $\operatorname{IoU}$ over all overlapping frames between track pairs and remove duplicates whose average $\operatorname{IoU}$ exceeds a threshold.

\paragraph{Post-processing.}
We remove tracks that are unlikely to be valid instances: (i) person-category tracks without any face detection, (ii) tracks whose area falls below a threshold, and (iii) a number of outliers/erroneous detections via a final manual pass.

\paragraph{Result.}
The resulting corpus contains long videos with time-aligned dense captions and multiple consistently tracked references (up to 15 per video). Consistent masks provide both reliable presence timestamps for each entity and enable mask-based data augmentation by sampling entity masks that lie outside the sampled training frames, thereby providing complex pose/lighting/appearance changes.
We visualize some of the annotation results obtained in~\cref{supp_fig:data_viz}.

\section{Multi-CFG}
\label{supp_sec:multi_cfg}
Due to presence of multiple input conditions in terms of reference text, images, time intervals as well as global, dense captions, we use multiple passes for Classifier-Free-Guidance~\cite{ho2022classifier}(CFG) for the different conditions. However, all combinations of the input conditions would be prohibitively expensive growing exponentially. We therefore group the reference text and images into a joint reference condition for dropping. We also group the global and dense captions as a joint text condition similar to~\cite{wu2025mint}. We do not drop/zero out reference time intervals as altering WeRoPE leads to undesirable patchy artifacts. As highlighted in~\cite{instructpix2pix,video_alchemist}, we obtain the multi-CFG equation as

\begin{align*}
\tilde{e}_{\theta}(z_t, c_{\text{ref}}, c_{\text{text}})
&= \quad e_{\theta}(z_t, c_{\text{ref}}, c_{\text{text}}) \\
&+ w_{\text{text}} \cdot \big(e_{\theta}(z_t, c_{\text{ref}}, c_{\text{text}}) - e_{\theta}(z_t, c_{\text{ref}}, \varnothing)\big) \\
&+ w_{\text{ref}} \cdot \big(e_{\theta}(z_t, c_{\text{ref}}, c_{\text{text}}) - e_{\theta}(z_t, \varnothing, c_{\text{text}})\big) \\
&+ w_{\text{both}} \cdot \big(e_{\theta}(z_t, c_{\text{ref}}, c_{\text{text}}) - e_{\theta}(z_t, \varnothing, \varnothing)\big).
\end{align*}
with $e_{\theta}(z_t, c_{\text{ref}}, c_{\text{text}})$ being the score estimation function with the reference and text conditioning, $c_{\text{ref}}$ and $c_{\text{text}}$ respectively at denoising timestep $t$. We set $w_{\text{text}}=8, w_{\text{ref}}=2, \ \text{and} \ w_{\text{both}}=3$. We perform denoising via rectified flow sampling for 40 steps at $288{\times}512$ resolution with time-shifting value of $5.66$.

\begin{figure*}[t]
  \centering
  \includegraphics[width=0.98\linewidth]{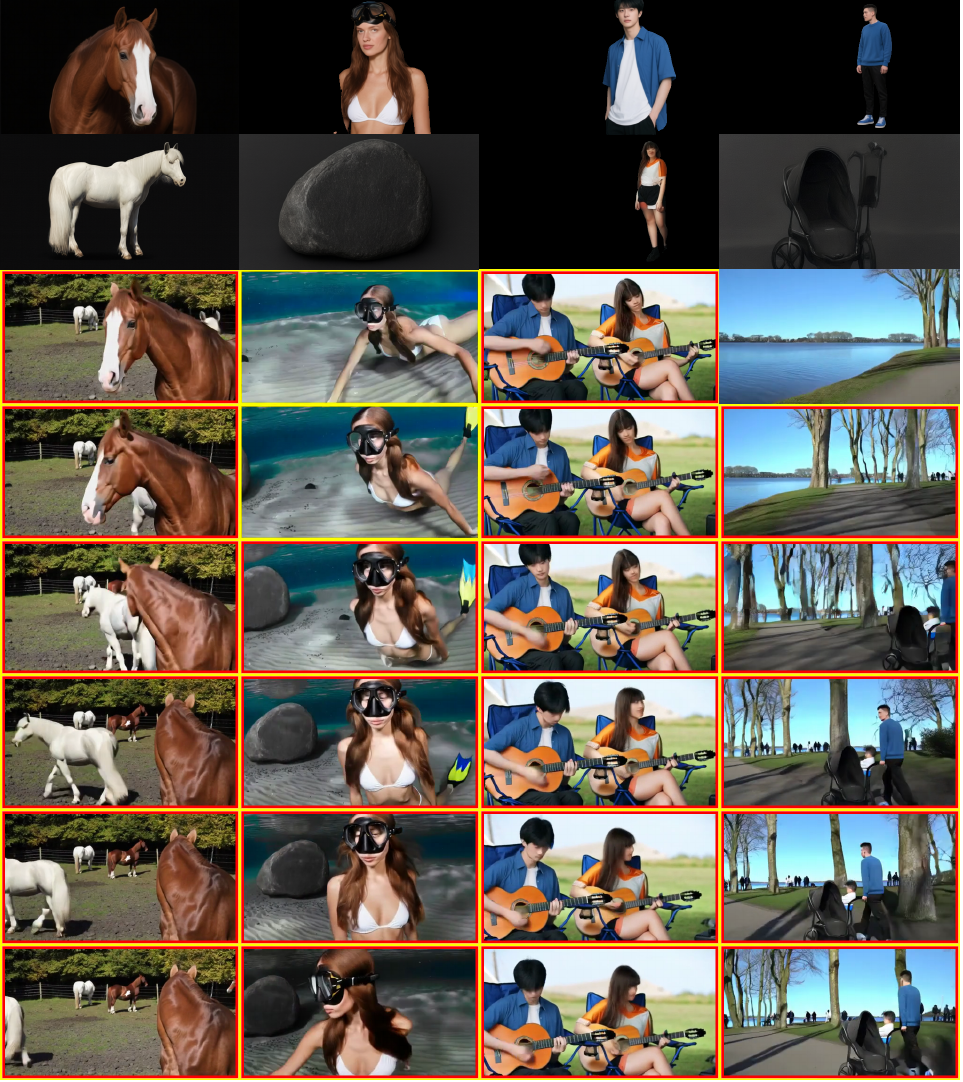}
  \caption{\textbf{Additional qualitative visualizations.} Yellow boxes indicate the expected occurence for the first reference and red boxes for the second. We obtain high fidelity generations which closely follow the input time interval.}
  \label{supp_fig:addn_qualitative_1}
  \vspace{-6pt}
\end{figure*}

\begin{figure*}[t]
  \centering
  \includegraphics[width=0.98\linewidth]{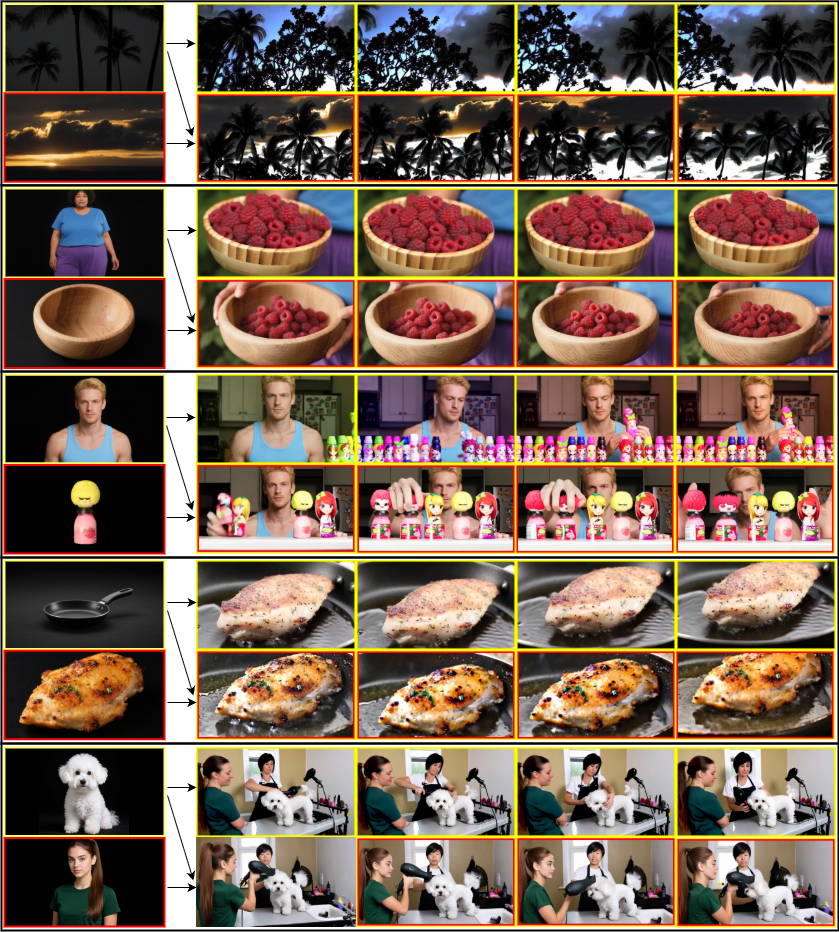}
  \caption{\textbf{Varying number of references .} We qualitatively visualize videos generated with a single reference as input or 2 references. We use the same set of text prompts as input for both generations. As expected, we see that including the second reference produces generations which are more consistent with the input preserving the identity appearance information.}
  \label{supp_fig:1_vs_2_ref}
  \vspace{-6pt}
\end{figure*}

\section{RoPE premilinaries}
\label{supp_sec:rope}

RoPE injects position by rotating each 2D feature subspace of a token with a phase that depends on its index~\cite{rope}.
Let the model dimension be $d$ (even) and define a bank of angular frequencies
\[
\boldsymbol{\omega}=\{\omega_i\}_{i=0}^{\frac{d}{2}-1},
\qquad
\omega_i = 10000^{-\,2i/d}.
\]
For a 1D index $n\!\in\!\mathbb{Z}$, the phase vector is
$\boldsymbol{\phi}(n)=\{\,\phi_i(n)=n\,\omega_i\,\}_{i}$, grouping real coordinates into complex pairs
$z_i = z_{2i}+ \mathrm{i} z_{2i+1}$, $i=0,\dots,\tfrac{d}{2}-1$.
RoPE applies a rotation (complex multiplication) independently to each pair:
\begin{equation}
\begin{aligned}
\langle \hat{\boldsymbol{q}}, \hat{\boldsymbol{k}} \rangle
&= \operatorname{Re}\!\Bigg[
   \sum_{i=0}^{\frac{d}{2}-1}
   q_i(k_i)^{*}
   \exp\!\Big\{\mathrm{i}\big(\phi_i(m)-\phi_i(n)\big)\Big\}
\Bigg] \\[-2pt]
&\qquad\text{(with } q_i=q_{2i}+\mathrm{i}q_{2i+1},\; k_i=k_{2i}+\mathrm{i}k_{2i+1}\text{).}
\end{aligned}
\end{equation}

\paragraph{3D extension.}
For video tokens with coordinates $m=(x,y,t)$, channel pairs are split across the three axes (e.g., $d_x+d_y+d_t=d$) and use axis-specific frequency banks
$\boldsymbol{\omega}^{(x)}, \boldsymbol{\omega}^{(y)}, \boldsymbol{\omega}^{(t)}$.
A convenient notation is to write the total phase for pair $i$ as the sum of axis phases,
\[
\phi_i(m) \;=\; x\,\omega^{(x)}_i + y\,\omega^{(y)}_i + t\,\omega^{(t)}_i,
\]
and apply the same rotation rule per pair.

\paragraph{Relative-position property.}
Let $\boldsymbol{q},\boldsymbol{k}\in\mathbb{R}^{d}$ be query/key at positions
$m$ and $n$, and let $q_i,k_i$ be their complex pairs.
After RoPE, the dot product depends only on the \emph{relative} position:
\[
\langle \hat{\boldsymbol{q}}, \hat{\boldsymbol{k}} \rangle
\;=\;
\operatorname{Re}\!\left[
  \sum_{i=0}^{\frac{d}{2}-1}
  q_i \,(k_i)^{*}\,
  e^{\mathrm{i}\,(\phi_i(m)-\phi_i(n))}
\right].
\]
In the common 1D case with $\phi_i(n)=n\,\omega_i$, this reduces to
\[
\operatorname{Re}\!\left[
  \sum_{i=0}^{\frac{d}{2}-1}
  \mathbf{q}_{[2i:2i+1]}\,\mathbf{k}^{*}_{[2i:2i+1]}\,
  e^{\mathrm{i}(m-n)\,\omega_i}
\right],
\]
where $\mathbf{q}_{[2i:2i+1]}=q_{2i}+\mathrm{i}q_{2i+1}$ and similarly for $\mathbf{k}$, and
$\operatorname{Re}[\cdot]$ denotes the real part.
This complex/rotation view makes clear that RoPE enforces a phase difference proportional to the relative displacement, yielding a natural inductive bias for long-range, relative attention.

\section{Additional visualizations}
\label{supp_sec:addn_qualitative}
In addition to the qualitative visualizations in the main paper, we provide further results on our benchmark videos. The supplementary material includes short video files demonstrating generations with single- and two-reference inputs, as well as frame-wise visualizations for multi-reference cases (\cref{supp_fig:addn_qualitative_1,supp_fig:1_vs_2_ref}). In the figures, yellow boxes mark the expected presence window of the first reference (from the input interval), and red boxes mark the second reference. As illustrated in \cref{supp_fig:addn_qualitative_1}, the model produces high-quality, reference-consistent samples that largely respect the specified timing (e.g., the rock, red box, in column~2; the man with stroller in column~4).~\cref{supp_fig:1_vs_2_ref} also shows generations with only the first reference, and both references with the same prompt. We see that providing the second reference produces generations consistent with the image in terms of its corresponding attributes showing that the model is able to produce consistent generations with the subject/reference condition.

Note that there are slight mismatches between interval inputs and when a reference actually appears in the generation due to a) the error in temporal downsampling for the latents (by a factor of 4) and also b) RoPE producing a gradual decay in attention score and not a sharp falloff as visualized in Fig. 3 of the main paper. This is however, necessary, for producing smooth generations with references gradually appearing in the video without abrupt unnatural transitions.

{
    \small
    \bibliographystyle{ieeenat_fullname}
    \bibliography{main}
}

\end{document}